\documentclass[10pt,twocolumn,letterpaper]{article}
\usepackage[pagenumbers]{cvpr}

\usepackage{graphicx}
\usepackage{caption}
\usepackage{multicol}
\usepackage{amsmath}
\usepackage{amssymb}
\usepackage{booktabs}
\usepackage{algorithm}
\usepackage{algpseudocode}
\renewcommand{\vec}[1]{\mathbf{#1}}

\usepackage{tabularx}
\usepackage{multirow}
\usepackage{amsmath}
\usepackage{amssymb}
\usepackage{graphicx}
\usepackage{xcolor}
\usepackage{colortbl}
\usepackage{stmaryrd}
\usepackage{amsthm}
\usepackage{bm}

\renewcommand{\vec}[1]{\mathbf{#1}}

\newcommand{\X}{\mathbf{X}}

\newcommand{\x}{\vec{x}}

\usepackage[pagebackref,breaklinks,colorlinks]{hyperref}

\usepackage[capitalize]{cleveref}
\crefname{section}{Sec.}{Secs.}
\Crefname{section}{Section}{Sections}
\Crefname{table}{Table}{Tables}
\crefname{table}{Tab.}{Tabs.}

\definecolor{cvprblue}{rgb}{0.21,0.49,0.74}

\def\confName{CVPR}
\def\confYear{2024}

\title{MarkovGen: Structured Prediction for Efficient Text-to-Image Generation}

\author{
Sadeep Jayasumana \qquad Daniel Glasner \qquad Srikumar Ramalingam \qquad Andreas Veit \\
Ayan Chakrabarti \qquad Sanjiv Kumar \vspace{0.1cm} \\
Google Research, New York \\
{\tt\small \{sadeep, dglasner, rsrikumar, aveit, ayanchakrab, sanjivk\}@google.com}
}

\begin{document}

\twocolumn[{%
\renewcommand\twocolumn[1][]{#1}%
\maketitle
\begin{center}
    \setcounter{figure}{0}
    \centering
    \captionsetup{type=figure}
    \vspace{-1.5em}
    \includegraphics[width=0.8\textwidth,trim={0.0cm 0.0cm 0cm 0},clip,page=1]{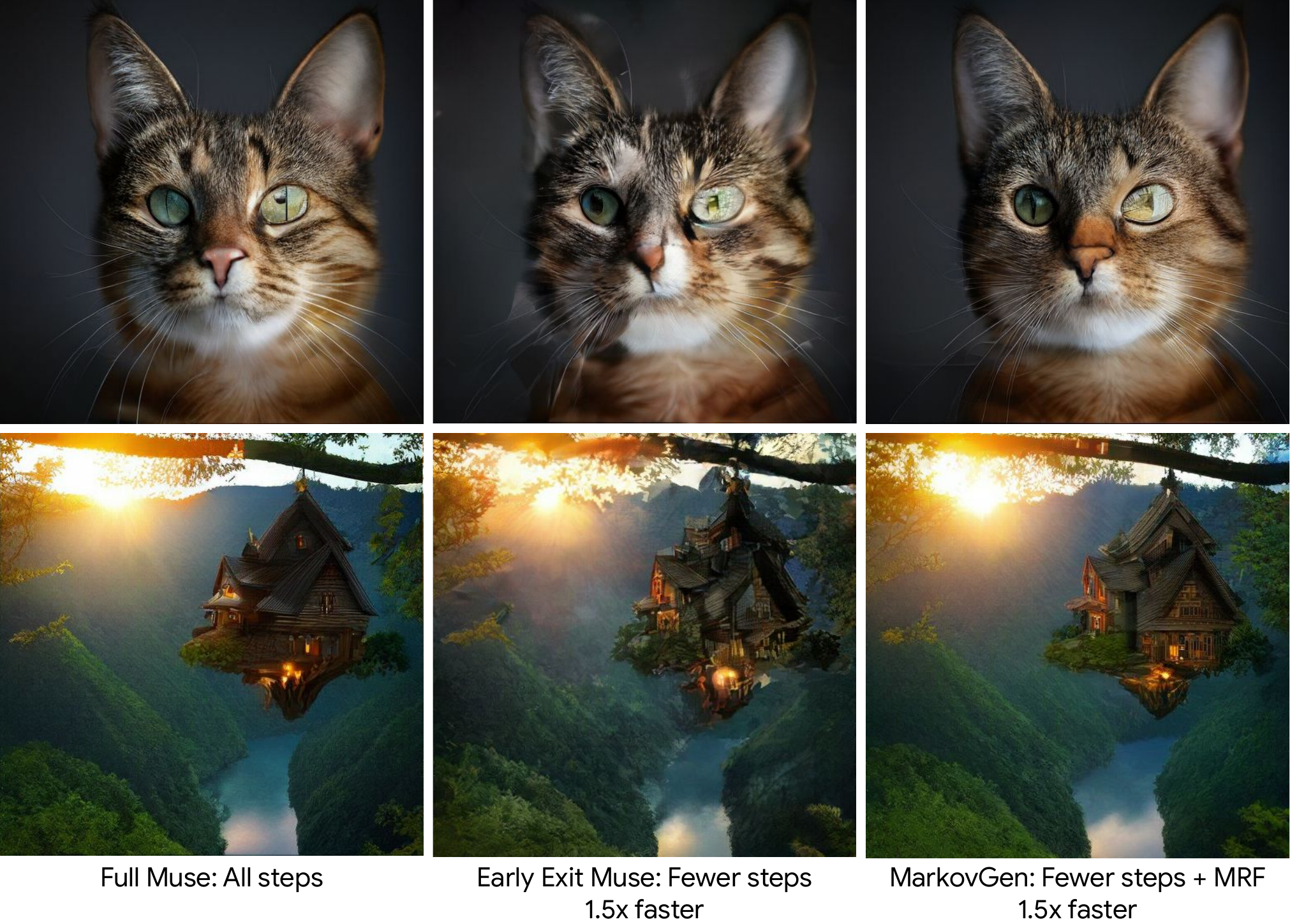}\vspace{-0.5em}%
    \caption{\em MarkovGen improves the speed and quality of token-based image generation models such as Muse, by reducing the number of sampling steps and replacing them with a light-weight Markov Random Field (MRF) model.}\label{fig:teaser}%
\end{center}%
}]

\begin{abstract}
Modern text-to-image generation models produce high-quality images that are both photorealistic and faithful to the text prompts. However, this quality comes at significant computational cost: nearly all of these models are iterative and require running sampling multiple times with large models. This iterative process is needed to ensure that different regions of the image are not only aligned with the text prompt, but also compatible with each other. In this work, we propose a light-weight approach to achieving this compatibility between different regions of an image, using a Markov Random Field (MRF) model. We demonstrate the effectiveness of this method on top of the latent token-based Muse text-to-image model. The MRF richly encodes the compatibility among image tokens at different spatial locations to improve quality and significantly reduce the required number of Muse sampling steps. Inference with the MRF is significantly cheaper, and its parameters can be quickly learned through back-propagation by modeling MRF inference as a differentiable neural-network layer. Our full model, MarkovGen, uses this proposed MRF model to both speed up Muse by $1.5\times$ and produce higher quality images by decreasing undesirable image artifacts.
\vspace{-0.5cm}
\end{abstract}

\section{Introduction}

Recent image-to-text models~\cite{saharia2022photorealistic,Rombach2022,yu2022scaling,ramesh2022hierarchical,Midjourney2022} are remarkably successful at producing high-quality, photo-realistic images that are faithful to the provided text prompts, and are poised to drive a new generation of tools for creativity and graphic design. However, the generation process with these models is iterative and computationally expensive, requiring multiple sampling steps through large models. For example, diffusion models~\cite{saharia2022photorealistic,Rombach2022} require multiple denoising steps to generate the final image, the Parti model~\cite{yu2022scaling} auto-regressively generates image tokens one at a time. While the recently proposed Muse model~\cite{chang2023Muse} generates multiple tokens at a time, it still requires a large number of sampling steps to arrive at the final image.

This iterative process is needed to ensure that different regions or patches of the images are not only aligned with the provided text prompt, but also \emph{compatible with each other}. Current text-to-image models achieve this spatial compatibility by repeatedly applying their full model multiple times on intermediate image predictions---a process that is computationally very expensive. In this paper, we demonstrate that a significantly lighter-weight approach can achieve the same compatibility.

\begin{figure}[t]
\centering
\includegraphics[width=0.3\textwidth]{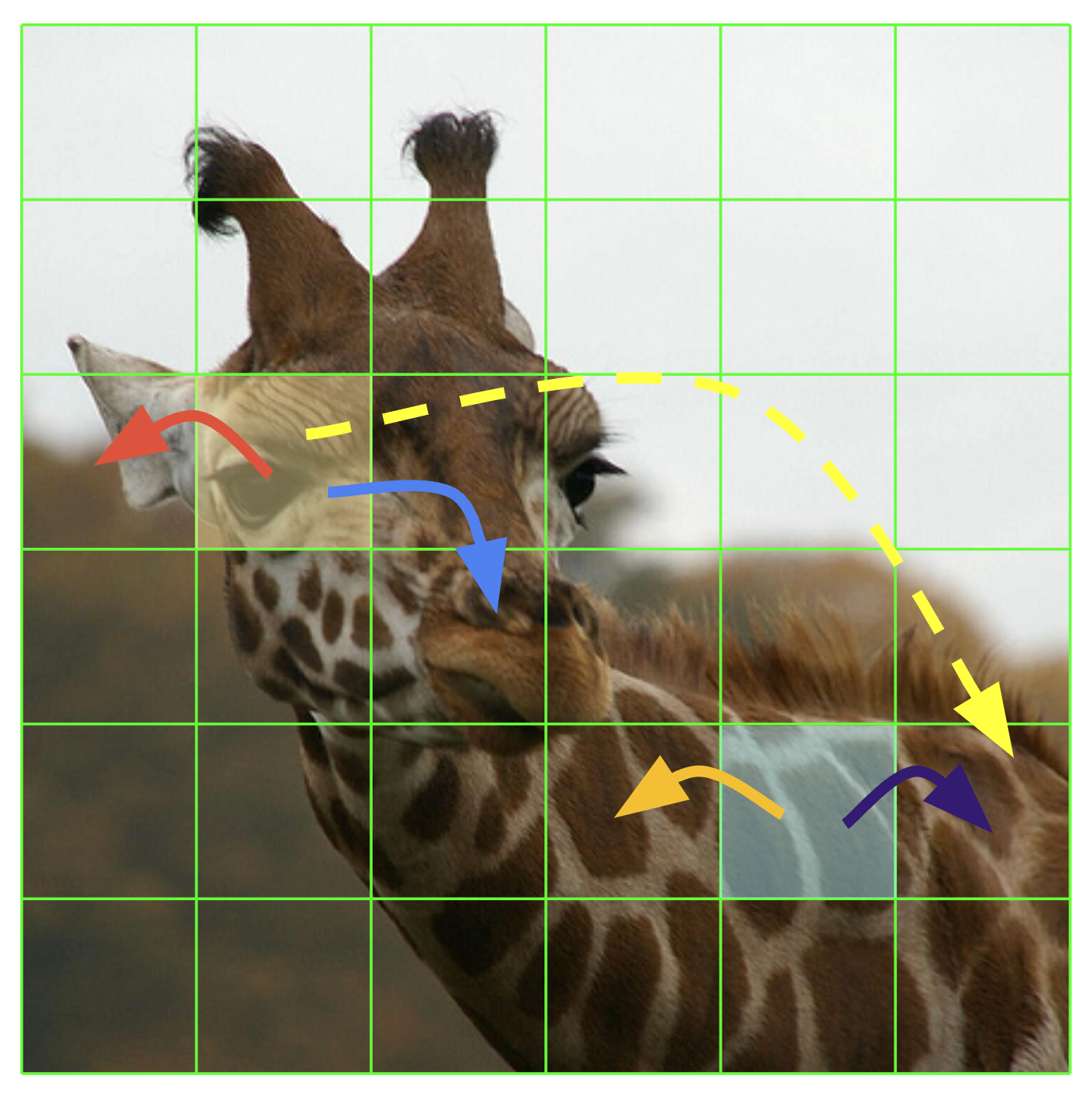}
\caption{{\it Benefits of encouraging token compatibility with an MRF model. During MRF inference, a confident token, such as the token representing the giraffe's eye, encourages the neighboring tokens to be compatible to represent other parts of a giraffe face such at ears and nose. Similarly, tokens representing the texture of the giraffe body can influence nearby tokens to encourage consistent patterns. Our formulation also supports long-range connections, such as the one shown with the dashed yellow line. }}
\label{fig:giraffe}
\vspace{-12pt}
\end{figure}

To this end, we propose a new \emph{structured prediction} approach that applies to image generation models operating in a discrete token space, such as the VQGAN token space~\cite{chang2022maskgit, esser2021taming, chang2023Muse}. These models generate images by first selecting tokens in a fixed-size token grid and later detokenizing them into an RGB image. Usual token-based image generation methods select tokens by \emph{independently} sampling from the probability distributions at different patch locations. In contrast, we model the whole image \emph{jointly} using a fully-connected Markov Random Field (MRF) that encodes compatibility between all pairs of tokens (image patches). The tokens at different patch locations are then determined based on this joint distribution. Consequently, as illustrated in Figure~\ref{fig:giraffe}, a confident token at one location can influence the selected tokens at other locations to enhance the overall compatibility of the token arrangement, and therefore the fidelity of the final image. We use mean-field inference~\cite{Koller2009, Krahenbuhl2011, Zheng2015} to solve this MRF, which also permits training the compatibility parameters of the model through back-propagation. During image generation with a trained model, the MRF inference comes at a negligible cost compared to the cost of large Transformer models used to predict the initial token probabilities.

To showcase the benefits of our MRF model, we introduce a new text-to-image model, MarkovGen, that can work in conjunction with the Muse model~\cite{chang2023Muse}. Muse uses a parallel decoding approach where all tokens of an image are predicted in parallel at each step. 
Muse has been shown to be much faster (around $3\times$ faster than the closest competitor) than other state-of-the-art image generation models such as DALL-E, Imagen, Parti, and Stable Diffusion, while producing similar or better quality images~\cite{chang2023Muse}.  Although Muse produces predictions for every patch simultaneously, single-shot parallel decoding leads to serious quality degradation in the generated images~\cite{chang2023Muse}. Muse solves this by embracing progressive parallel decoding, where a small incremental number of high confidence tokens are fixed after each iteration. We show that by learning the compatibility of the tokens and applying the MRF inference after limited number of sampling steps with the Muse model, we achieve significant quality and efficiency gains over the Muse's full iterative approach (see Figure~\ref{fig:teaser}).

Reducing the latency and improving the quality of Muse, one of the fastest text-to-image models, will have important practical implications for real-world deployments. The success of our MRF formulation in modeling spatial and label relationships of image tokens opens up the future possibility of refining predictions of other token-based methods such as Parti~\cite{yu2022scaling} and discrete-diffusion models~\cite{gu2022vector} with MRFs. %

In summary, our contributions are:
\begin{itemize}

\item We propose an MRF model, a type of probabilistic graphical model, that can predict a globally compatible set of image tokens by explicitly modeling spatial and token label relationships. To the best of our knowledge, this is the first work to exploit MRFs to improve the efficiency and quality of text-to-image models.

\item Our MarkovGen model, where we replace the last few steps of Muse with a learned MRF layer, leads to a $1.5\times$ speedup as well as improved quality results, as demonstrated by human evaluation and FID distances. 

\item We show that the MRF model parameters can be trained in just a few hours, allowing us to quickly combine the MRF model with pre-trained Muse models to reap efficiency and quality gains. 

\end{itemize}

\section{Related Work}

\noindent{\bf Text-to-Image Generation:} In recent years, papers such as \cite{chang2022maskgit,chang2023Muse,saharia2022photorealistic,ramesh2022hierarchical,Rombach2022,yu2022scaling,zhang2023adding,singer2022makeavideo,villegas2022phenaki,ge2022long,kumari2023multiconcept} have proposed a diverse variety of methods to generate high-quality images given a text prompt as input. We discuss some of the most relevant approaches below.

Many text-to-image models~\cite{saharia2022photorealistic,ramesh2022hierarchical,Rombach2022,Nichol2022,Midjourney2022,gafni2022makeascene} use denoising diffusion probabilistic models (DDPM)~\cite{Ho2020} to generate images, where the model is invoked successively to ``denoise'' previous intermediate versions and progressively refine the image output. While in theory, we need infinitely small and many denoising steps, only a few hundreds of steps are used in practice~\cite{pmlr-v37-sohl-dickstein15}. Progressive distillation algorithms are being developed to cut down the number of steps~\cite{salimans2022progressive}. Most of these models directly operate on and produce pixel intensities, \cite{Rombach2022,zhang2023adding} are variants that operate on a lower-dimensional latent representation.

In contrast, the Parti~\cite{yu2022scaling}, DALL-E~\cite{ramesh2021zero}, and Muse~\cite{chang2023Muse} models generate images in a space of discrete token representation. They use a VQGAN~\cite{esser2021taming} model, derived from VQVAE~\cite{Oord2017}, to represent non-overlapping image patches with tokens---with values from a discrete vocabulary---and cast the image generation task as that of generating image tokens. The Parti and DALL-E models approach token generation with auto-regressive modeling, generating tokens one at a time in sequence, where each token is generated conditioned on the text input and all previously generated tokens.

The Muse model~\cite{chang2023Muse}, on the other hand, is trained to take the text prompt and any already generated image tokens as input, and make predictions for all remaining image tokens simultaneously. In particular, it is trained as a BERT-style~\cite{devlin2018bert} encoder model operating on a masked set of image tokens (with tokens not already generated being masked), with cross-attention to an encoding of the text prompt input. To generate an image, the model is invoked in multiple sampling steps, with all image tokens being masked in the first step. At each step, the Muse model makes predictions for all masked tokens. A subset of these predictions are selected and added to the set of fixed and non-masked tokens, which are then used as conditioning input for subsequent invocations till the all tokens have been fixed.  Similar to Muse, Paella~\cite{rampas2022fast} and Cogview2~\cite{ding2022cogview2} also exploit progressive parallel decoding to achieve speedup. A similar approach to parallel decoding for text was introduced by \cite{Ghazvininejad2019}.

Like many other text-to-image generation models, Muse first generates a low-resolution version of the target image, and then conditions on this low-resolution image to generate the high-resolution version. It uses a similar architecture and sampling approach for the high-resolution generation stage, except in this case, the low-resolution image tokens are provided as additional conditioning input.

For the selected tokens at each sampling stage of Muse, the token values are determined independently for each token from the predicted per-token distributions. Our structured prediction approach, in contrast, considers compatibility between the values of different tokens, and by doing so, is able to improve the quality and reduce the number of sampling steps required---in both the low- and high-resolution stages.

Many of the text-to-image algorithms are also being extended to develop algorithms to handle other conditional inputs~\cite{zhang2023adding}, and text-to-video generation~\cite{singer2022makeavideo,villegas2022phenaki,ge2022long}. 

\noindent{\bf Structured Prediction:} Markov and Conditional random fields (CRF)s have a long
history of being used in computer vision for diverse applications such as stereo, segmentation, and image reconstruction~\cite{szeliskimrf}. These MRF and CRF models have typically been used to enforce smoothness constraints, i.e., that semantic labels, pixel intensities, stereo depths, etc. at nearby locations are similar. In neural network-based methods too, they have been a useful post-processing step~\cite{chen2015semantic} to yield smooth consistent results.

While early MRF and CRF models considered edges only among immediate pixel neighbors on the image plane, \cite{Krahenbuhl2011} introduced ``fully-connected'' CRF models that had far longer range connections, and showed that the energy for these models could effectively be minimized using mean-field inference. Using this fully-connected formulation, \cite{Zheng2015} proposed back-propagating through the mean-field inference steps to jointly train a CRF model with a CNN network to achieve better semantic image segmentation.

In this work, we use an MRF formulation to achieve consistency in predicted image tokens in the context of text-to-image generation, and like \cite{Zheng2015}, we also use a fully connected MRF model and learn its parameters by back-propagation. However, in our case, the MRF is defined over tokenized patches, the label space corresponds to the vocabulary of a VQGAN~\cite{esser2021taming} and the MRF enforces consistency between different token values rather than smoothness. 

It is worth mentioning here that CRFs have recently also been proposed to improve text generation~\cite{Sun2019,Su2021}. Like our case, these methods also use a Transformer model to generate ``unaries'' that are then provided as input to a CRF model. However, these methods  consider edges only between neighboring tokens, and since text sequences are one-dimensional, are able to use chain decoding techniques (like beam search) for inference. In contrast, our method reasons with a two-dimensional MRF model with edges between all pairs of patches in the image.

\section{Structured Token Prediction}

\begin{figure}[!t]
\centering
\includegraphics[width=0.47\textwidth]{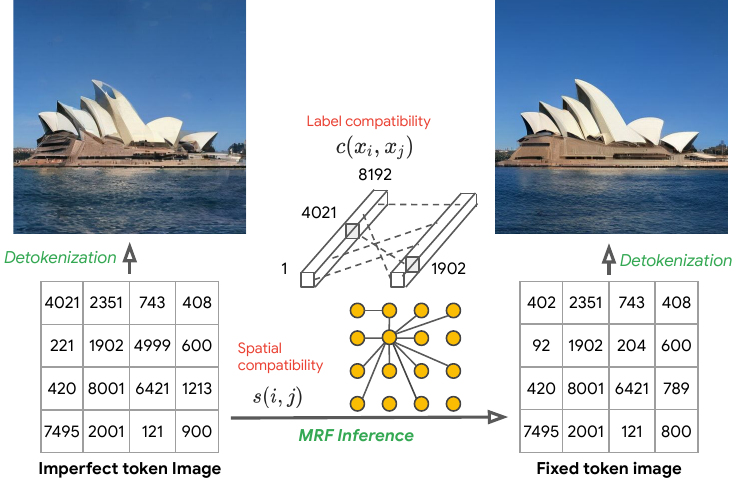}
\vspace{-3pt}
\caption{{\it Given individual token probabilities from an underlying Transformer-based image generation backbone, the MRF improves image quality by utilizing learned spatial and label compatibility relations in the latent token space.}}
\label{fig:mrf_mod}
\vspace{-10pt}
\end{figure}

In this section, we introduce our MRF formulation for structured token prediction. In token-based image generation, a neural network (often a Transformer model) makes predictions to generate a fixed size ($16 \times 16$, for example) \emph{token image} containing token labels. This token image is then sent through a \emph{detokenizer} to generate an RGB image~\cite{esser2021taming}. 

Consider a common vocabulary size of $V=8192$. For a full sized $16 \times 16$ image, there are $8192^{256} = 6.7 \times 10^{1002}$ different arrangements of tokens, many of which will represent some kind of ``garbage'' images that lie outside the manifold of photorealistic images. Intuitively, a structured prediction mechanism that accounts for the compatibility of token arrangements could significantly reduce this massive search space of token arrangements and make the token prediction models more efficient. 

We propose a probabilistic graphical model for this structured prediction task. Specifically, we formulate finding the token arrangement as maximum a posteriori (MAP) inference of an MRF model, as described in the following. The high-level idea is illustrated in Figure~\ref{fig:mrf_mod}.

Let $i \in \{1, 2, \dots, n\}$ denote the location indices of the token image, arranged in row-major order. Let $\mathcal{L} = \{l_1, l_2, \dots, l_V\}$ be the token labels, which are used to index each element in the codebook of V tokens.  For a $16 \times 16$ token image with vocabulary size $8192$, we have $n = 256$ and $V = 8192$. Define a random variable $X_i \in \mathcal{L}$ for each $i = 1, 2, \dots, n$ to hold the token assignment for the $i^\text{th}$ location. The collection of these random variables $\vec{X} = [X_1, X_2, \dots, X_n]$ then forms a random field, where the value of one variable depends on that of the others. We can then model the probability of an assignment to this random field (and therefore a token arrangement on the grid) with the Gibbs measure:
\begin{equation}
    P(\X = \x) = \frac{1}{Z}\exp(-E(\x)),
\end{equation}
where $\x \in \mathcal{L}^n$ is a given token arrangement and $Z = \sum_{\x} \exp(-E(\x))$ is the partition function. The “energy” $E(\vec{x})$ of an assignment $\x = \{x_1, x_2, \dots, x_n\}$ is modeled with two components: the \emph{unary} component $u_i(.)$ and the \emph{pairwise} component $p_{ij}(., .)$:

\begin{align}
    E(\vec{x}) = \sum_{i=1}^n u_i(x_i) + \sum_{i=1}^n\sum_{j=1}^{n} p_{ij}(x_i, x_j).
\label{eq:energy_function}
\end{align}

\begin{algorithm}[t]
\caption{\it The MRF Inference Algorithm}\label{alg:cap}
\begin{algorithmic}
\State $Q_i(k) \gets \operatorname{softmax}(f_i(k))$, $\forall (i, k)$
\For {\texttt{num\_iterations}}
    \State $Q_i(k) \gets \sum_{j=1}^n \mathbf{W^{s}}_{ij} Q_j(k)$, $\forall (i, k)$ %
    \State $Q_i(k) \gets \sum_{k'=1}^V \mathbf{W^{c}}_{kk'} Q_i(k')$, $\forall (i, k)$ %
    \State $Q_i(k) \gets Q_i(k) + {f}_i(k)$, $\forall (i, k)$ %
    \State $Q_i(k) \gets \operatorname{softmax}(Q_i)(k)$, $\forall (i, k)$ %
\EndFor
\State \textbf{return} $Q$
\end{algorithmic}
\end{algorithm}

The unary component captures the confidence of the neural network prediction model, such as a Transformer model, for a given token and a location. Therefore, given a condition $y$, such as a text prompt or pre-fixed tokens, if the neural network's predicted logit value for location $i$ and label $x_i$ is $f_i(x_i, y)$, we set:

\begin{align}
    u_i(x_i) = -f_i(x_i, y).
\end{align}
Note that we use negative logits because the energy function is in the log domain and a high energy corresponds to a low probability. We drop conditioning on $y$ hereafter to keep the notation uncluttered. Also note that our MRF formulation is not conditioned on $y$.

The pairwise component, $p_{ij}(x_i, x_j)$, captures the compatibility of the label $x_i$ assigned to the location $i$ and the label $x_j$ assigned to the location $j$. It encodes the notion that while some pairs of tokens are highly compatible with each other and can appear in the same image, other pairs are highly incompatible. For example, a token representing an eye of a giraffe is more likely to appear next to a token representing a different part of a giraffe face, than a token representing something completely different like a part of car wheel. We factorize this pairwise compatibility into two parts: the spatial similarity $s(i, j)$ between the locations $i$ and $j$ (for example, if $i$ and $j$ are close to each other in the 2D token image, they will be strongly related) and the label compatibility $c(x_i, x_j)$ between the tokens $x_i$ and $x_j$ (for example, highly compatible tokens are able to coexist with each other). We therefore have:

\begin{align}
    p_{ij}(x_i, x_j) = -c(x_i, x_j) s(i, j).
\end{align}

In classic MRFs, the pairwise interactions exist only between neighboring pixels. In contrast, for increased flexibility, we allow interactions between all pairs of locations, similar to the fully-connected CRFs in the image segmentation setting~\cite{Krahenbuhl2011, Zheng2015}. However, there are a number of important differences in our formulation compared to the fully-connected CRFs in image segmentation: in the latter, spatial similarity $s(i, j)$ is derived conditioned on the input image (hence the name \emph{conditional} random fields), using Gaussian potentials in spatial and bilateral domains. This Gaussian assumption is crucial for the tractability of their models since the image segmentation CRFs work in a large image grid: in practical implementations pixels that are far away by more than a few standard deviations of the Gaussian kernel are considered not connected~\cite{adams2010}. In contrast, we make our graphical model truly fully-connected and learn $s(i, j)$ with backpropagation without fixing them to be Gaussian. Furthermore, the CRFs in image segmentation can assume a Potts model for label compatibility because assigning the same label to nearby pixels generally improves the smoothness of the segmentation. In our application, on the other hand, it is not straightforward to assign semantic meanings to tokens and Potts model does not intuitively makes sense since the same token at similar locations does not increase the meaningfulness of a token assignment. We therefore learn the pairwise connections $p_{ij}(., .)$, completely from data without using any priors or heuristics. Thus, our MRF formulation has two learnable weight matrices: $\mathbf{W^s}$, with $\mathbf{W^s}_{ij} := s(i, j)$ and $\mathbf{W^c}$, with $\mathbf{W^c}_{kk'} := c(k, k')$.

Given our probabilistic graphical model, finding the final token arrangement amounts to finding the assignment $\x$ that maximizes $P(\X=\x)$. This can be done efficiently via mean-field inference, where we approximate $P(\X) \approx Q(\X) := \prod_i Q_i(X_i)$, with $Q_i(.)$ being the marginal distribution for $X_i$. The distribution $Q(\X)$ is then iteratively refined to minimize the KL divergence between $P$ and $Q$. We refer the reader to \cite{Koller2009} and \cite{Krahenbuhl2011} for more details on the derivations. The resulting inference algorithm is summarized in Algorithm~\ref{alg:cap}. Note that all operations of this algorithm can be implemented via simple matrix multiplication and other common operations such as $\operatorname{softmax(.)}$, which are readily available in any deep learning library. Importantly, the cost our MRF inference is negligible compared to prediction with a large Transformer model.
\section{MarkovGen}

\begin{table}[t]
    \centering
    \begin{tabular}{l r}
        \toprule
        Model & Time (ms) \\
        \midrule
        Muse base (single step) & 10.40  \\
        Muse super-resolution (single step)  & 24.00 \\
        MRF inference on base  &  0.29 \\
        MRF inference on super-resolution & 0.29 \\
        T5-XXL inference & 0.30\\
        Detokenizer & 0.15\\
        \midrule
        Muse  & 442.05 \\
        MarkovGen (ours) & 281.03 \\
        \bottomrule
    \end{tabular}
    \caption{\it Average inference times for different components of the MarkovGen models on a TPUv4 device. The {\sc MRF} inference is almost free compared to the costs of the Muse Transformer models. Furthermore, MRF inference is independent of the image resolutions (rows 3 and 4). We make Muse inference $1.5\times$ faster by introducing the MRF model.}
    \label{tab:inference_times}
\end{table}

We now demonstrate the benefits of the proposed MRF model by using it to speed up the state-of-the-art Muse image generation model~\cite{chang2023Muse}. We achieve this speed-up by replacing the last few sampling steps of Muse with MRF inference. Specifically, we let Muse execute the first few steps and then use our extremely lightweight MRF inference to fast-forward the remaining steps. This model, dubbed MarkovGen, improves the speed of image generation by $1.5\times$ while simultaneously also improving quality.

\begin{figure*}[!t]
\centering
\includegraphics[width=1.0\textwidth,trim={0cm 0cm 0cm 0},clip,page=1]{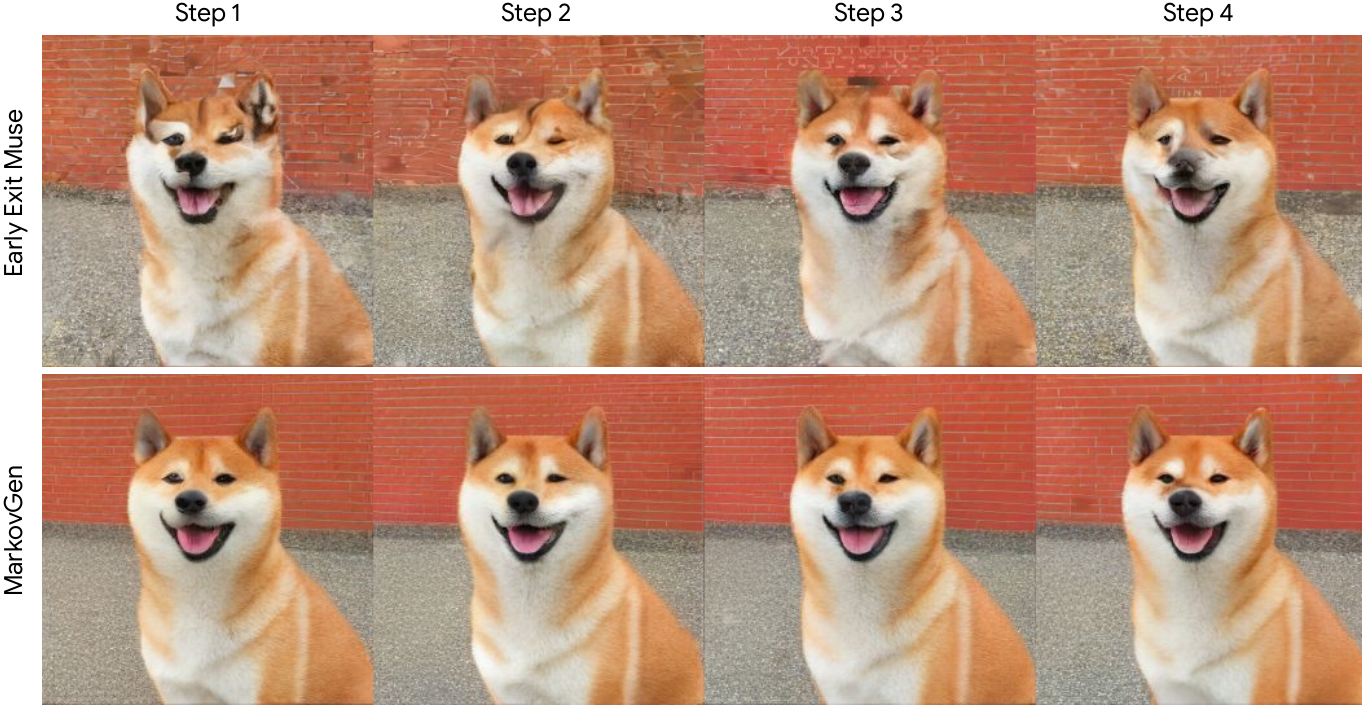}
\caption{\it The first four steps of the Muse super-resolution model without (top) and with (bottom) the application of the MarkovGen MRF model. Note that the MRF fixes complex object structures such as the dog's face as well as texture-inconsistencies in areas such as the brick wall. MarkovGen generates good looking high quality images starting from the first step.}
\label{fig:sr_progression}
\vspace{-10pt}
\end{figure*}

The Muse model works in the discrete VQGAN token space~\cite{esser2021taming}, which is gradually emerging as a centerpiece of many text-to-image generation algorithms.
Muse generates images by first performing a series of inference steps with the base model to predict a small grid of $16 \times 16$ image tokens, conditioned on text embeddings generated by a T5-XXL~\cite{Raffel2020} text encoder. This is followed by a few steps of the super-resolution (SR) model to predict a larger grid of $32 \times 32$ image tokens by conditioning on both the text embeddings and the tokens generated by the base model. 
Due to the larger set of tokens, a single iteration of the SR transformer model is substantially more computationally expensive than that of the base model. 
We exploit this multi-scale approach to speed up inference efficiency by using more steps in the base model, followed by much fewer steps with the SR model. Once SR tokens are generated, the VQGAN~\cite{esser2021taming} detokenizer is used to render the image in pixel space. 

The goal of MarkovGen is to fast-forward the later part of the Muse model and replacing it with the MRF outlined in the previous section. To achieve this, we train the MRF to match the final predictions of the Muse model, given the output at an intermediate step. By fast-forwarding after the step $k$, out of a total of $n$, we instantly save $(n-k)/n \times 100\%$ of the Muse model's inference time. This is because the inference time of the MRF is negligible compared to that of the Muse steps as shown in Table~\ref{tab:inference_times}. The same strategy is used for both the base model and the SR model, to achieve an overall boost of $1.5\times$ in inference speed.

Muse determines the token values independently at each sampling stage, and our structured prediction, enforces the learned compatibility relations jointly on the different token values, and thereby leading to improved quality as demonstrated in the experiments. 
\section{Experiments}

In this section, we show that MarkovGen achieves both faster inference and improved image quality compared to Muse, which it uses as its backbone. Muse by itself has already been shown to be much faster than other state-of-the-art text-to-image models such as Dall-E, Dall-E 2, Parti, Imagen, and Stable Diffusion~\cite{chang2023Muse}, outperforming the second-fastest method by a factor of approximately $3 \times$ (Table 3 of~\cite{chang2023Muse}). Furthermore, as evidenced by Table 1 \& 2 of \cite{chang2023Muse}, Muse achieves better quality results compared to these methods, as measured with the FID scores. Human evaluation results for image quality in \cite{chang2023Muse} showed that humans preferred Muse outputs for $70.6\%$ prompts while Stable Diffusion was preferred for only $25.4\%$. Since Muse is already shown to outperform other state-of-the-art methods in terms of both speed and quality, we focus on comparing our results to that of Muse.

\begin{figure*}[!t]
\centering
\includegraphics[width=0.72\textwidth]{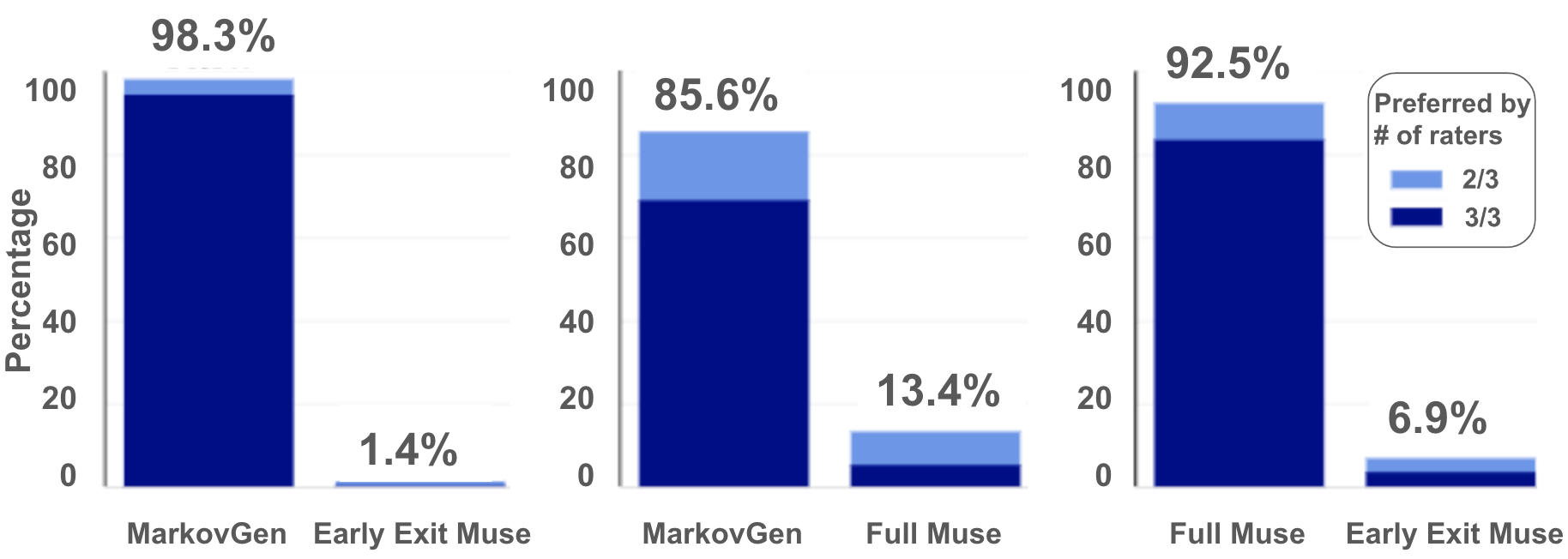}
\vspace{-5pt}
\caption{\it Percentage of prompts for which human raters prefer images by a given model in a side-by-side comparison. We observe that human raters strongly prefer the images generated by MarkovGen over those of both early exit Muse (left) and even the more expensive and slower full Muse model (center).}
\label{fig:human_eval}
\vspace{-8pt}
\end{figure*}

\vspace{0.5cm}
\noindent{\bf Model and Dataset:}
We use a Muse model with approximately 1.7B parameters, trained on the WebLI dataset~\cite{Pali2022}. This model was generously made available to us by the authors of the Muse paper. We refer the reader to~\cite{chang2023Muse} for more details on the architecture and the training setup of Muse. The same WebLI dataset was used to train the MRF model.

\vspace{0.5cm}
\noindent{\bf MRF Training:} 
We train two MRF models, one for fast-forwarding the base and one for SR model respectively. Each model contains two weight matrices for spatial and label compatibilities: $256 \times 256$ $\mathbf{W^s_{base}}$  and $8192 \times 8192$ $\mathbf{W^c_{base}}$ for thr base, and $1024 \times 1024$ $\mathbf{W^s_{SR}}$ and $8192 \times 8192$ $\mathbf{W^c_{SR}}$ for the SR model. All MRF weights are trained with back-propagation and gradient descent, with the ADAM optimizer.
We use a two-stage approach for MRF training. First, we pre-train the MRF model using a self-supervised masked-token prediction loss~\cite{devlin2018bert, chang2023Muse}. Specifically, we obtain VQGAN tokens for an image, randomly mask $20\%$ of them and train the MRF model to predict the masked tokens using the categorical cross-entropy loss. Second, we fine-tune the MRF model to imitate the last $n-k$ steps of the Muse model: Given the output of the Muse model after the $k$th iteration, the spatial and label compatibility matrices are learned such that the MRF inference matches the final predictions of the Muse model after $n$ iterations using the KL divergence loss. 
Both base and SR MRF models are trained in the same manner. Both MRF models completes training in just a few hours on TPUv4 chips.

\vspace{0.5cm}
\noindent{\bf Experimental Setup:}  The base model operates on a $16 \times 16$ token grid with 24 sampling steps to produce $256 \times 256$ images. The SR model works on a $32 \times 32$ token grid and produces $512 \times 512$ images in 8 additional steps. MarkovGen uses both the base and SR MRF models to trade with the base and SR sampling steps of the Muse model, respectively.  We apply the base MRF after step 20 of the base Muse model, and the SR MRF after 3 steps of the SR Muse model, cutting down 4 and 5 steps respectively. This results in a speedup of $1.5 \times$ for MarkovGen compared to Muse.

In our experiments we compare 3 different models: (A) full Muse, (B) MarkovGen, and (C) an early exit Muse model that also stops Muse iterations early to have comparable speed to MarkovGen, but without the application of the MarkovGen MRF model.

\begin{table}[t]
    \centering
    \begin{tabular}{cc}
        \toprule
        Model           & FID \\
        \midrule
        Early Exit Muse base (18 iters)     &  14.37   \\
        Full Muse base (24 iters)  & 13.13 \\
        MarkovGen (18 iters) &  12.28  \\
        \bottomrule
    \end{tabular}
    \vspace{-5pt}
    \caption{\it Quantitative evaluation of FID scores on the MS-COCO~\cite{lin2014microsoft} dataset for $256 \times 256$ image resolution. MarkovGen outperforms both the Early Exit as well as the full Muse model.}
    \label{tab:fids}
    \vspace{-5pt}
\end{table}

\begin{figure*}[t]
\centering
\includegraphics[width=1.0\textwidth,trim={0cm 0cm 0cm 0},clip,page=1]{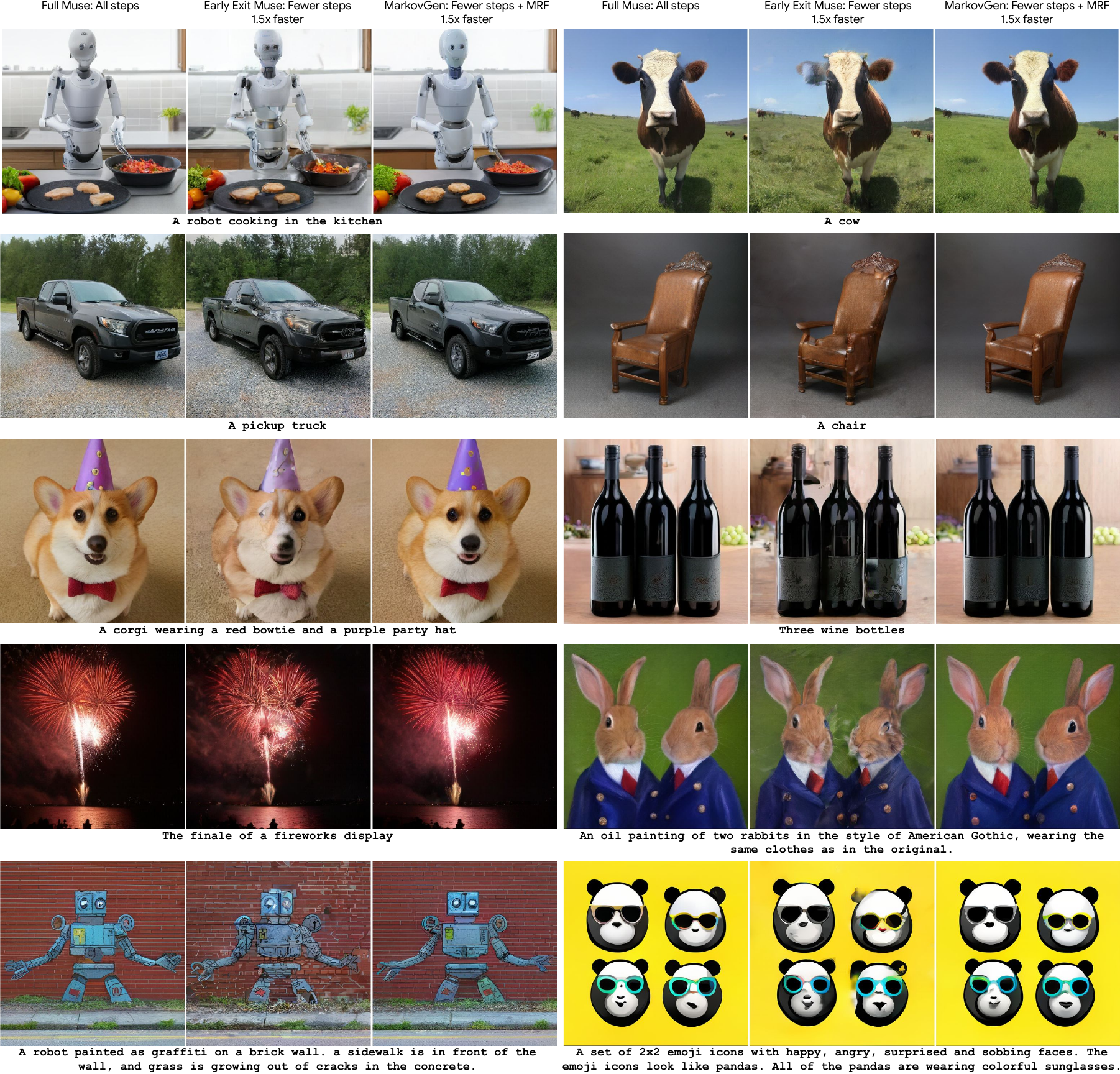}
\vspace{-15pt}
\caption{\it Within each set of three, MarkovGen (right) speeds up Muse (left) by $1.5 \times$ and improves image quality. A similar speed up by only reducing the step count with early exit Muse (middle) results in a significant loss of quality.}
\label{fig:sets_of_three_1}
\end{figure*}

\vspace{0.5cm}
\noindent{\bf Qualitative Evaluation:} 

In Figure~\ref{fig:sr_progression} we study the progression of the generated images during the invocation of the Muse SR model. At each step, we show the output of early exit Muse (top) and the improved result after the application of the MarkovGen MRF (bottom). The results show that while the Muse model slowly improves result quality, MarkovGen provides high quality results already after the first step. We observe that MarkovGen with just 3 SR steps already consistently produces images comparable or better than the full Muse results after the total of 8 SR steps. Figure~\ref{fig:sets_of_three_1} demonstrates this using a series of Parti Prompts~\cite{yu2022scaling}, where we compare the results of the full Muse model (left), early exit Muse (middle) and, our MarkovGen model with a $1.5\times$ speed-up (right). We observe that the model is able to produce a wide variety of images ranging from artistic to natural images.

\vspace{0.5cm}
\noindent{\bf Quantitative Evaluation:} 
Figure~\ref{fig:human_eval} summarizes the results of our human evaluation study. Using the 1633 prompts from the Parti promts dataset, we generated three images with full Muse, early exit Muse, and our proposed MarkovGen model respectively. To allow  raters to focus on image quality, we use the same random seed across models to ensure that image content and degree of alignment to the prompt are the same across the generated images. 
Asked to evaluate which image is of higher quality, we present human raters with two generated images side-by-side. Raters are given the option of choosing either image or that they are indifferent. All image pairs are evaluated by 3 independent raters that were hired through a high-quality crowd computing platform. The raters and the authors of this paper were anonymous to each other. For each pairwise comparison, we consider an image to be of higher quality if it is selected by at least 2 raters. 

From the results we observe that human raters strongly prefer the images generated by MarkovGen over those of both early exit Muse (left) and even the more expensive and slower full Muse model (center). We also compared early exit Muse to full Muse (right) and verified that human raters can clearly identify the quality improvement achieved by the last stages of the Muse model. This result demonstrates that MarkovGen not only achieves a drastic speed-up of 1.5x over Muse, but also significantly improves image quality.

In addition to human evaluation, Table~\ref{tab:fids} shows single-shot FID scores on the MS-COCO dataset~\cite{lin2014microsoft}. We use the base Muse model for this evaluation. Again, in line with the human evaluation, we observe that MarkovGen achieves better results than full Muse, which in turn outperforms early exit Muse.

\vspace{0.5cm}
\section{Conclusion}

The proposed MarkovGen model showed a significant inference speed-up of $1.5\times$ and a clear quality gain over Muse by fast-forwarding the last few steps of Muse model with MRF inference. The MRF model achieves this by learning the spatial and token label compatibility relationships in the discrete VQGAN token space. Our MRF model can be trained in just a few hours, allowing us to use it in conjunction with pre-trained Muse models, to observe almost immediate improvements. 

While providing clear benefits over independent per-patch token selection, our current MRF model does not yet utilize the provided text prompt, with text guidance coming solely through the unaries. An interesting direction of future work would be to make the spatial and token compatibility weights be dependent on the text prompt, allowing the MRF (or in this case, the CRF) to adapt to text input.

Another direction of future work lies in training the Muse model itself jointly with the MRF layers, so as to ensure that the unaries produced by Muse are optimal for use with MRF-based decoding.

\section*{Acknowledgment}
We would like to thank Apurv Suman, Dilip Krishnan, Jarred Barber, Huiwen Chang, and 
Jason Baldridge for their valuable feedback. We thank the authors of Muse~\cite{chang2023Muse} for generously providing us with the code and the models.

{\small
\bibliographystyle{ieee_fullname}
\bibliography{refs}
}

\begin{figure*}[t]
\centering
\includegraphics[width=0.9\textwidth,clip,page=1]{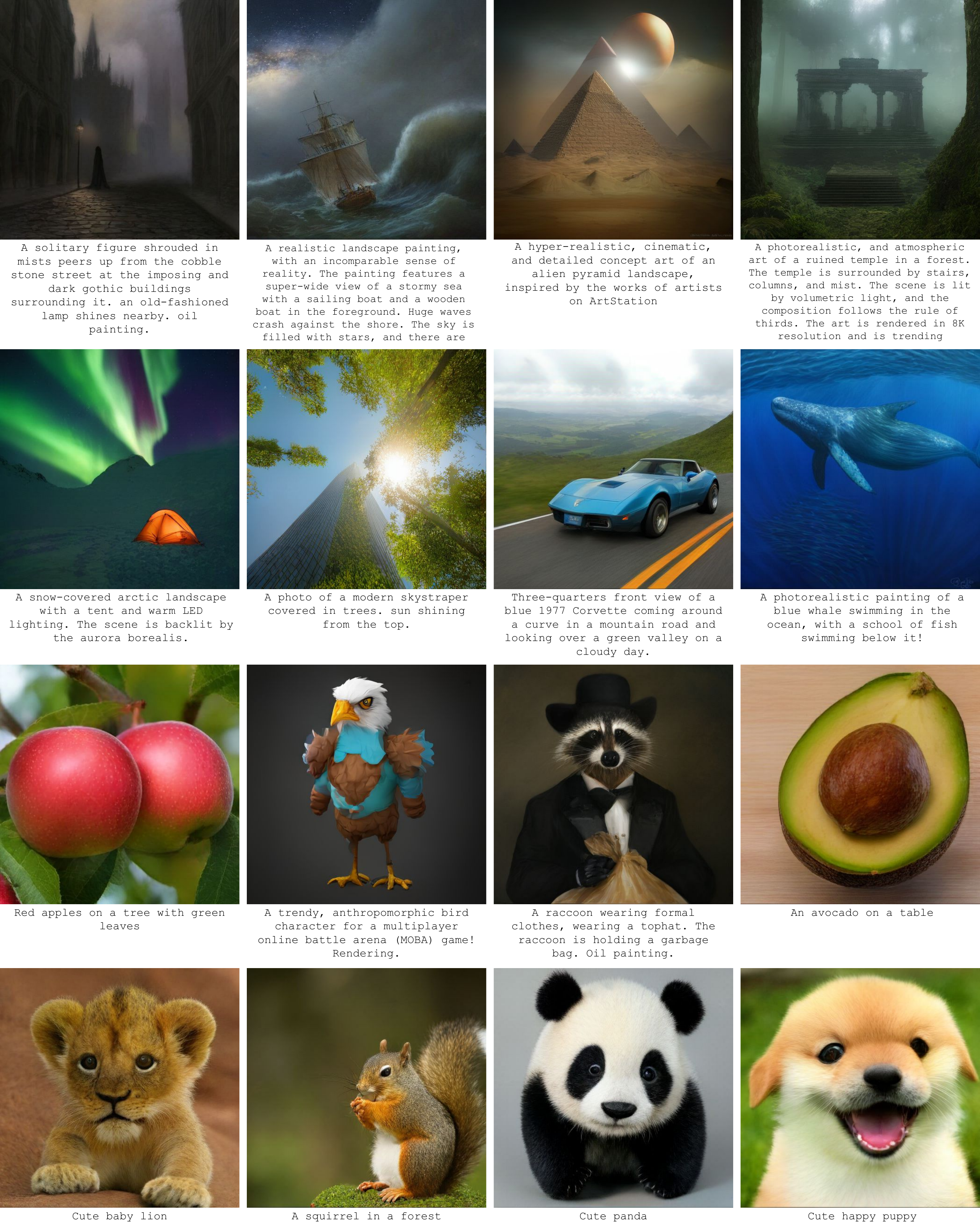}
\caption{\it Images generated by MarkovGen for a wide variety of prompts ranging from artistic to natural images.}
\label{fig:sr_gallery}
\end{figure*}

\newpage
\appendix
\twocolumn[{%
\renewcommand\twocolumn[1][]{#1}%

\section{Appendix}

\subsection{Additional Qualitative Results}
In this section we present additional qualitative results. In Figure~\ref{fig:mrf_results_super} we compare the results of the Early Exit Muse model to the results of the MarkovGen model. This direct comparison helps to illustrate the impact of the {\sc MRF} model to image quality. We can see that the MarkovGen model with shows clear imporvements to complex object structures as well as visual artifacts and textures. In Figure~\ref{fig:sr_gallery} we show generations of the MarkovGen model for a broad variety of text prompts demonstrating the ability of the proposed MarkovGen model to produce high quality images from a broad set of text prompts.

\begin{center}
    \centering
    \captionsetup{type=figure}
    \includegraphics[width=1.0\textwidth,trim={0.0cm 0.0cm 0cm 0},clip,page=1]{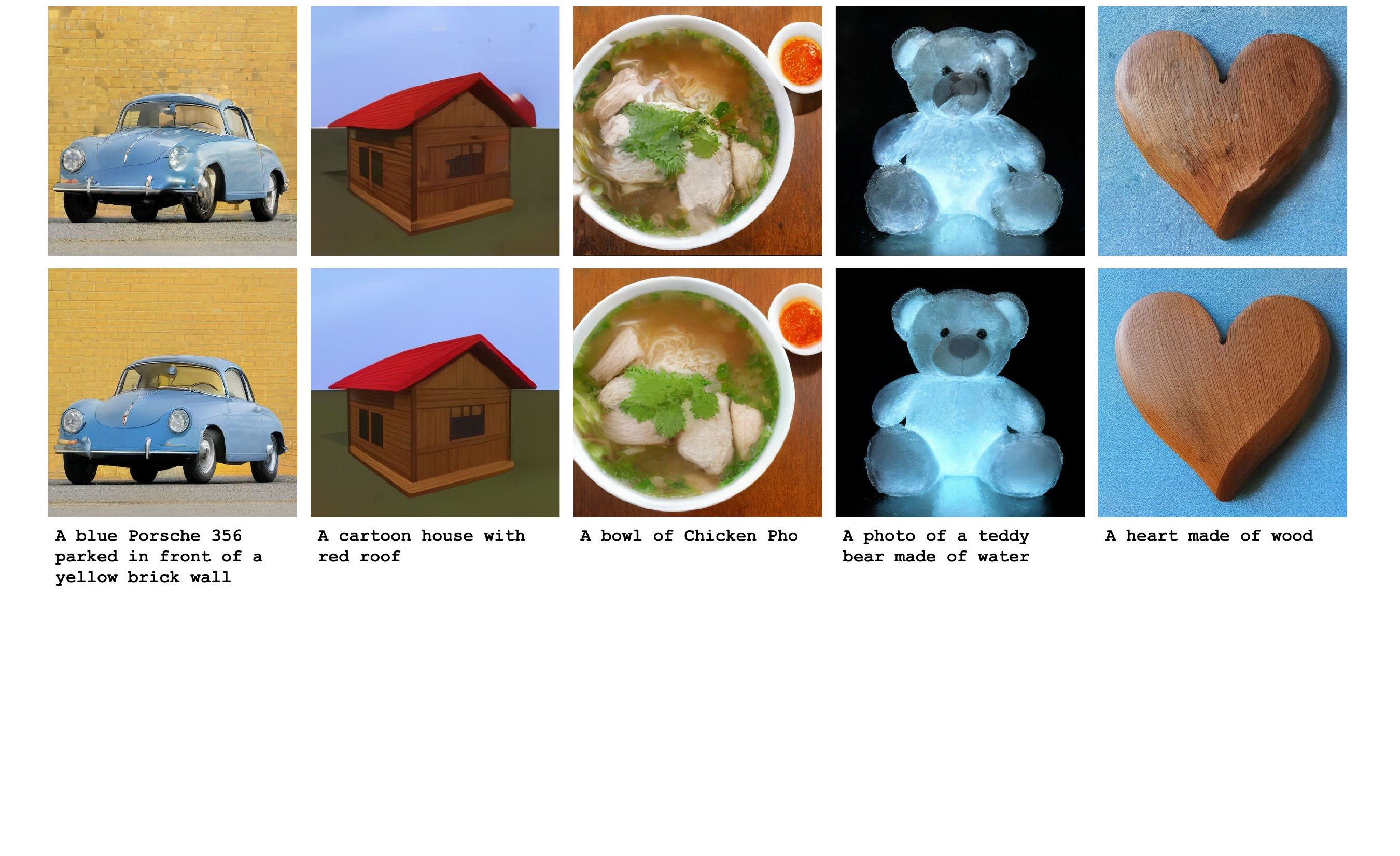}\vspace{-85pt}%
    \caption{\it Example generations of the Early Exit Muse super-resolution model running for 3 (out of 8) steps (top) and the MarkovGen model after the application of the {\sc MRF} model (bottom). We observe a significant reduction in visual artifacts, e.g., in the brick wall behind the car. We further see key improvements to complex object structures such as the blue car and the teddy bear's face.}\label{fig:mrf_results_super}%
\end{center}%
}]

\end{document}